\documentclass[runningheads]{llncs}

\usepackage{accv}

\usepackage{accvabbrv}

\usepackage{graphicx}
\usepackage{booktabs}
\usepackage[ruled,vlined]{algorithm2e}
\usepackage{multirow}
\usepackage[accsupp]{axessibility}  

\usepackage{hyperref}

\usepackage{orcidlink}

\begin{document}

\title{Shot Segmentation Based on Von Neumann Entropy for Key Frame Extraction} 

\titlerunning{Von Neumann Entropy for Key Frame Extraction}

\author{Xueqing Zhang\inst{1}\orcidlink{0000-0003-4531-6482} \and
Di Fu\inst{2} \and Naihao Liu \inst{1}\orcidlink{0000-0002-2609-7408}}

\authorrunning{X. Zhang et al.}

\institute{School of Information and Communications Engineering, Xi'an Jiaotong University, Xi’an, Shaanxi, China, \email{xue\_qing\_zhang@163.com}, \email{naihao\_liu@mail.xjtu.edu.cn}
\and ByteDance, Beijing, China, \email{fudi.01@bytedance.com}}

\maketitle

\begin{abstract}
    Video key frame extraction is important in various fields, such as video summary, retrieval, and compression. Therefore, we suggest a video key frame extraction algorithm based on shot segmentation using Von Neumann entropy. The segmentation of shots is achieved through the computation of Von Neumann entropy of the similarity matrix among frames within the video sequence. The initial frame of each shot is selected as key frames, which combines the temporal sequence information of frames. The experimental results show the extracted key frames can fully and accurately represent the original video content while minimizing the number of repeated frames. 
  \keywords{key frame extraction \and Von Neumann entropy \and shot segmentation}
\end{abstract}

\section{Introduction}
\label{sec:intro}

With the rapid development of the Internet, videos have become the predominant forms of content on the Internet~\cite{Wang2015Robust} especially many short video platforms like TikTok are rising rapidly. Therefore, video summary~\cite{Apostolidis2021Video}, retrieval~\cite{Valentin2020Multi} and compression~\cite{Lu2021Compression} are becoming increasingly crucial. Key frame extraction is important in advancing these fields as it significantly reduces the consumption of computing resources and time. For video summaries, extracting key frames conveys the main content, saving time for viewers and facilitating efficient browsing. Additionally, Video storage space and transmission bandwidth can be reduced by saving and transmitting only the differences between key and subsequent frames. Therefore, an effective key frame extraction algorithm is needed. The key frame extraction algorithm should contain the content of a video using as few frames as possible.

Recently, several key frame extraction methods have been proposed including curve simplification~\cite{Halit2011Multiscale}, motion analysis\cite{Mo2021Motion}, clustering~\cite{Nasreen2015Clustering, Janwe2016VideoKE} and based on convolutional neural networks (CNNs)~\cite{Savran2021neural, Mo_2023_CVPR}. The curve simplification method utilizes certain image features to represent the data sequence as the track curve space in high-dimensional features. Nevertheless, this algorithm solely focuses on the local characteristics of the image, thus inadequately capturing its overall global characteristics. As a result, it often leads to redundancy in extracting key frames. The motion analysis method is primarily suitable for extracting key frames from moving objects, but it lacks effectiveness in extracting key frames related to dynamic changes in the background. The key frame extraction algorithm based on clustering typically clusters video frames by considering the global or local features of each frame. This approach aims to obtain representative frames from various categories. However, this algorithm requires determining the number of clusters in advance, and the adaptive method cannot efficiently obtain the cluster center either. Furthermore, an additional limitation of clustering algorithms is their disregard for the temporal sequence information of frames, which presents a significant drawback in the field of key frame extraction. The algorithm based on CNNs has high computational complexity and faces challenges in selecting the training dataset. To overcome these drawbacks, we propose a key frame extraction algorithm based on shot segmentation. The shot segmentation is achieved through the computation of Von Neumann entropy applied to the similarity matrix among frames. The initial frame of each shot is selected as the key frame of the video. This approach does not require prior knowledge of the exact number of key frames, thus overcoming the limitations of previous methodologies. Furthermore, the proposed algorithm integrates temporal information, which is a crucial consideration in key frame extraction. The algorithm has a computational complexity of \(O(N^2)\), indicating short processing time even when dealing with long input videos. The algorithm proposed achieves the effect results in the evaluation indexes: effective information rate, and redundancy rate. The extracted key frames can fully and accurately represent the original video content while minimizing redundant frames.

The main contributions of this study can be summarized as follows.
\begin{itemize}
    \item The key frames are obtained by segmenting video shots using Von Neumann entropy, which integrates temporal information among frames.
    \item The number of key frames can be obtained by analyzing the change of entropy, which does not require prior knowledge of the number of key frames.
    \item The algorithm has a computational complexity of \(O(N^2)\).
\end{itemize}

The rest of this study is organized as follows. In~\cref{sec:theoretical_basis}, we present the theoretical knowledge utilized in our algorithm. The detailed process of our approach is introduced in~\cref{sec:method}. The dataset, evaluation indicators, and experimental results are presented in~\cref{sec:experiments}. Finally, the conclusion is drawn in~\cref{sec:conclusion}.

\section{Theoretical Basis}
\label{sec:theoretical_basis}

\subsection{The Similarity of Frames}\label{ssec:similarity_frames}
Recently, several methods have been proposed for calculating image similarity, including the cosine similarity~\cite{XIA2015Learning}, histogram statistical~\cite{Strelkov2008histogram}, structure similarity index measure (SSIM)~\cite{Alain2010SSIM}, and hash algorithms~\cite{Wang2014hash}. The hash algorithm includes the mean hash algorithm, the difference hash algorithm, and the perceptual hash algorithm~\cite{MCKEOWN2023Hamming}. In the cosine similarity algorithm, the initial step involves converting the pixel matrix of an image into a normalized one-dimensional vector. Subsequently, the cosine distance between two vectors is computed to represent the similarity between the images. The algorithm weakens the influence of image texture, lighting, and shading, and cannot accurately represent the similarity between images. The histogram statistical algorithm only considers the color information of the image and does not consider the spatial position information of the image. Hence, the algorithm is unable to accurately capture the similarity of images. SSIM assesses image similarity based on brightness, contrast, and structure. However, the algorithm is not suitable for analyzing long videos due to its complexity.

The hash algorithm creates a unique "fingerprint" for each image and measures similarity by comparing these fingerprints. The mean hash is calculated by finding the average gray-scale value of the resized image and then comparing the pixels. The difference hash calculates the hash value by comparing the difference of gray-scale pixels. The perceptual hash uses discrete cosine transform (DCT)~\cite{Ahmed1974Discrete} to convert the gray-scale image from the pixel domain to the frequency domain, and the hash value is derived by comparing the values of the DCT matrix. In the frequency domain of an image, low frequency represents the fundamental gray level, corresponding to the color information, while high frequency reflects the edges and details. According to the theory of the human visual system~\cite{Ding2018}, the human eye is more sensitive to changes in smooth regions (low frequency) but less sensitive to changes in textured regions (high frequency). After performing DCT on an image, the majority of the frequency coefficients are concentrated within a relatively small range, mainly in the low-frequency region. This aligns with the principles of human vision. In addition, the perceptual hash algorithm has a positive impact on the adjusted size and angle of the image, so we choose perceptual hash to calculate the similarity.

\subsection{Von Neumann Entropy}\label{ssec:von_neumann_entropy}
The concept of entropy originated in physics as a measure of disorder in a thermodynamic system. In information theory, entropy is a measure of uncertainty. In Shannon entropy, higher entropy allows for greater transmission of information, while lower entropy limits the amount of information that can be transmitted~\cite{Alfr1961Entropy}. The frames within one shot are more similar and contain less information compared to frames from other shots. The experiment demonstrates that the entropy is significantly higher when adding frames from other shots. In our algorithm, key frames are extracted by dividing the video shots. Therefore, we can utilize entropy as a measure for video shot segmentation.

In Shannon entropy, the entropy of random variable X is defined as:
\begin{equation}
\label{eq:shannon_entropy}
H(X) = E[I(X)] = E[-ln(P(X))],
\end{equation}
where I is the amount of information, E is the expectation, and P is the probability distribution. If the random variable is taken from a finite sample, the entropy formula can be expressed as:
\begin{equation}
\label{eq:entropy_formula}
H(X) = \sum_iP(x_i)I(x_i)=-\sum_iP(x_i)log_bP(x_i),
\end{equation}
where \(x_i\) denotes random variables, P represents the probability distribution, I stands for the amount of information, and b represents the base of the logarithm, typically 2, e, or 10. In addition, Shannon entropy requires that \(\sum_iP(x_i)=1\)~\cite{Alfr1961Entropy}, which means the sum of the probabilities of the variables should be equal to 1. In our algorithm, the similarity matrix does not meet the required condition. Hence, the entropy of the similarity matrix cannot be calculated using information entropy.

The Von Neumann entropy is a quantum extension of Shannon entropy in information theory. For a given matrix \(\rho\), the Von Neumann entropy is defined as~\cite{Bengtsson2006Geometry,Zachos2007classical}:
\begin{equation}
\label{eq:VonNeumann_entropy}
H(\rho) = -\sum_q\lambda_qln\lambda_q,
\end{equation}
where \(\rho\) is the matrix, \(\lambda\) is the eigenvalue of the matrix, and q is the number of eigenvalues of the matrix. Therefore, the entropy of the similarity matrix can be calculated by determining the eigenvalues, and the algorithm's computational complexity is \(O(n^3)\).

\subsection{Beam Search}\label{ssec:beam_search}
The optimal segmentation minimizes the sum of entropy when dividing the video into shots. When we use the exhaustive search to segment the video, the computational complexity is \(O(n^n)\) which is terrible. The exhaustive search algorithm can achieve global optimization, but its computational complexity is too high, making it unsuitable for our task. The greedy search algorithm only keeps the current optimal result at each step, but it cannot guarantee that the final search results will be globally optimal.

Beam search is an improvement over greedy search. It expands the search space, but it is much smaller than the exhaustive search. Beam search is a heuristic graph search. During each step of the depth expansion, lower-quality nodes are eliminated, while higher-quality nodes are retained.
Beam search involves a hyperparameter known as the beam size, typically represented as K. This algorithm preserves the top K optimal results at each step and chooses the K results from the final step as the ultimate result. Greedy search can be seen as a special case of beam search with a beam size of 1. In our algorithm, we preserve the previous K segmentation that minimizes the sum of entropy. Finally, we choose the segmentation that minimizes the sum of entropy.

\section{Approach}\label{sec:method}
\begin{figure*}[!t]
\centering
\includegraphics[width=0.9\linewidth]{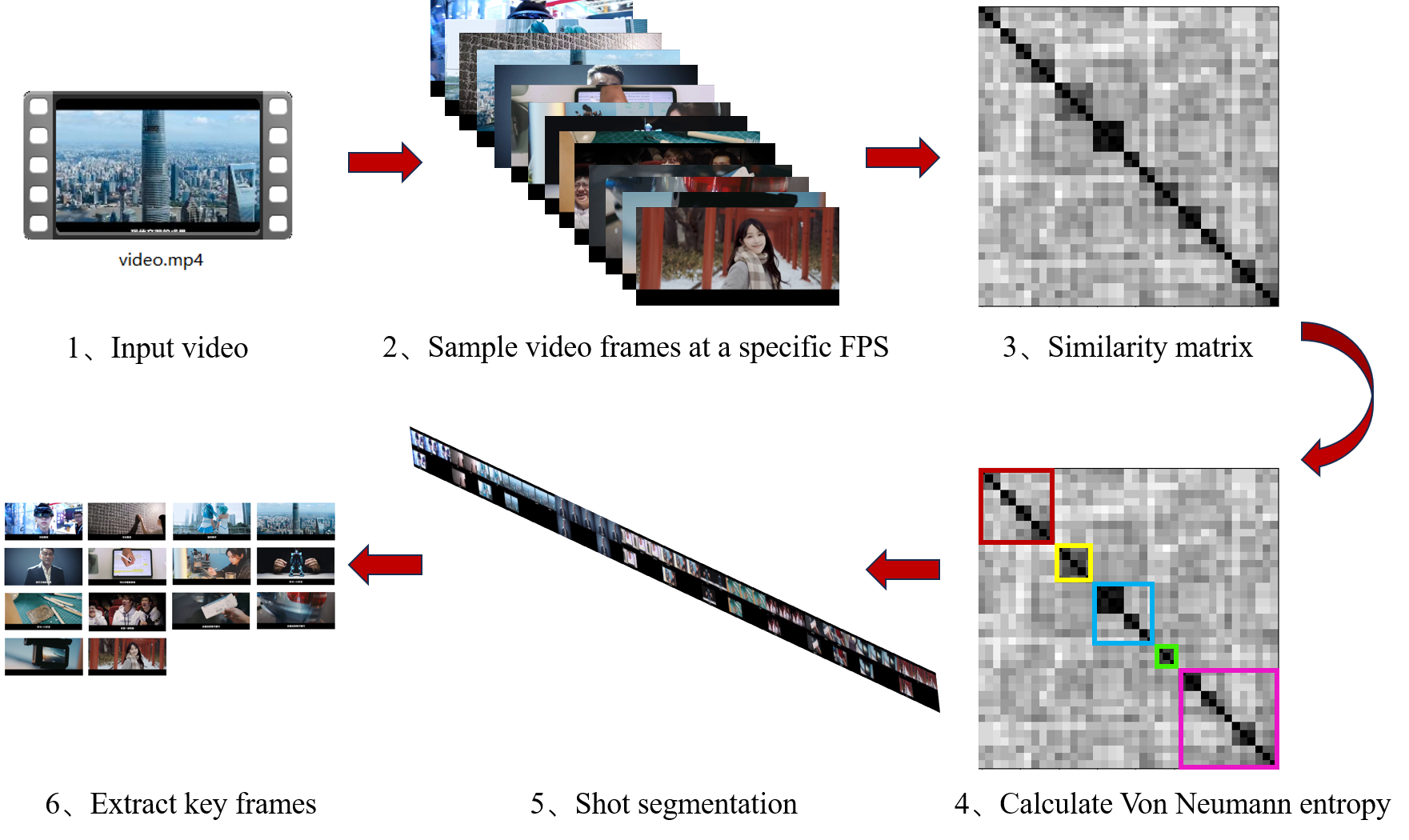}
\caption{The procedure of our approach.}
\label{fig:frame}
\end{figure*}

The procedure of our approach is illustrated in~\cref{fig:frame}. First, the video frames are sampled at specific frames per second (FPS) to reduce redundancies in the video data. Then the similarity matrix of sampled frames is calculated using the perceptual hash. After obtaining the similarity matrix, the Von Neumann entropy of the similarity matrix is computed. Then the video shots are segmented based on the Von Neumann entropy. Finally, we select the first frame of each shot as key frames. In addition, we compare the proposed algorithm with density peak clustering (DPC) using two datasets: the Open Video dataset and the dataset from TikTok. The proposed algorithm performs well in terms of effective information rate, and redundancy rate, as shown in the numerical experiments~\cref{ssec:experimental_results}. The extracted key frames fully and accurately represent the original video content while minimizing redundant frames.

\subsection{Similarity Matrix}\label{ssec:similarity_martix}

The video frames are sampled at a certain FPS, resulting in N sampled frames. Then the sampled frame is converted into the gray-scale and its DCT matrix is calculated. The \(8\times8\) matrix in the upper left of the DCT matrix is selected, and each value of the \(8\times8\) matrix is compared with the mean value of the \(8\times8\) matrix. If the value is equal to or larger than the mean value, mark this position as 1 bit; otherwise, mark it as 0 bit, resulting in a hash matrix. Comparing the hash matrices of two images to calculate their hamming distance D:
\begin{equation}
\label{eq:hamming_distance}
D = \sum_{i=1}^8\sum_{j=1}^8|h_{ij}^1-h_{ij}^2|,
\end{equation}
where \(h_{ij}^1\) and \(h_{ij}^2\) are the hash bits of two hash matrices at the same position. The similarity S between the two images is:
\begin{equation}
\label{eq:similarity}
S = 1-\frac{D}{64}.
\end{equation}

\begin{figure}[!t]
\centering
\includegraphics[width=0.4\linewidth]{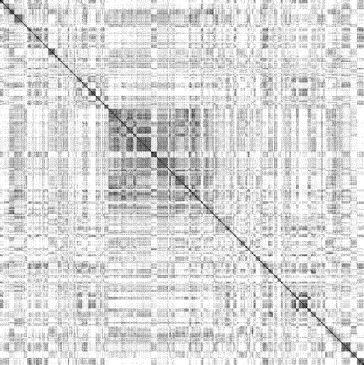}
\caption{We use a gray-scale image to represent the similarity matrix. The color closer to black indicates higher similarity between images, while the color closer to white indicates lower similarity.}
\label{fig:sim}
\end{figure}

Since the similarity matrix is diagonally symmetric, we only need to calculate half of the similarity between the sampled frames.~\Cref{fig:sim} shows a gray-scale image of a similarity matrix.

\subsection{Von Neumann Entropy Matrix}\label{ssec:von_neumann_entropy_matrix}

The entropy of the similarity matrix is calculated using Von Neumann entropy, which involves finding the eigenvalues of the similarity matrix. The similarity matrix of a long video leads to a significant increase in time consumption due to its large size. It is easier to calculate the sum of the eigenvalues of the matrix rather than calculating all of the eigenvalues. In this situation, we only need to compute the trace of the matrix. We utilize the \(O(n^2)\) computational complexity algorithm proposed by Thomas P. Wihler \etal~\cite{wihler2014computing} and Craig Gidney~\cite{Craig2016Computing} to calculate the entropy of the similarity matrix.
Thomas P. Wihler \etal suggest approximating \(xlnx\) with a polynomial, Craig Gidney suggests we can utilize the trace of powers of the matrix to sum the results obtained by applying the polynomial function to each eigenvalue. The similarity matrix has orthogonal eigenvectors, so we adopt Craig Gidney'algorithm to calculate the entropy. Taking the matrix to a power elevates its eigenvalues to the same power, while the trace gives the sum of these powered eigenvalues. Therefore, this algorithm only requires calculating the trace of the similarity matrix. As a result, the computational complexity is reduced from \(O(n^3)\) to \(O(n^2)\).

The derivatives of \(f(x)=xlnx\) are calculated as
\begin{equation}
    \label{eq:derivatives}
    \begin{aligned}
        f^0(x) & = xlnx, \\
        f^1(x) & = 1+lnx, \\
        f^2(x) & = x^{-1}, \\
        f^3(x) & = -x^{-2}, \\
        f^4(x) & = 2x^{-3}, \\
        & \cdots \\
        f^k(x) & = (-1)^k(k-2)!x^{1-k}.
    \end{aligned}
\end{equation}

The derivatives may not exist at \(x=0\), but they do at \(x=1\), allowing us to create a Taylor series from that point.
\begin{equation}
    \label{eq:Taylor}
    \begin{aligned}
        xlnx & = \sum_{k=0}^\infty\frac{(x-1)^k}{k!}f^k(1), \\
        & =(1ln1)\frac{(x-1)^0}{0!}+(1+ln1)\frac{(x-1)^1}{1!} \\ & +\sum_{k=2}^\infty\frac{(x-1)^k}{k!}(-1)^k1^{(1-k)}(k-2)!, \\
        & =x-1+\sum_{k=2}^\infty\frac{(1-x)^k}{k(k-1).}
    \end{aligned}
\end{equation}

We can estimate this series by truncating it at a specific index c:
\begin{equation}
    \label{eq:von_new_entropy}
    \begin{aligned}
        xlnx & =x-1+\sum_{k=2}^\infty\frac{(1-x)^k}{k(k-1)}, \\
        & =x-1+\sum_{k=2}^{c-1}\frac{(1-x)^k}{k(k-1)}+\sum_{k=c}^\infty\frac{(1-x)^k}{k(k-1)}, \\
        & \approx x-1+\sum_{k=2}^{c-1}\frac{(1-x)^k}{k(k-1)}+\sum_{k=c}^\infty\frac{(1-x)^c}{k(k-1)}, \\
        & =x-1+\sum_{k=2}^{c-1}\frac{(1-x)^k}{k(k-1)}+\frac{(1-x)^c}{c-1}.
    \end{aligned}
\end{equation}

The derivation above reduces the computational complexity from \(O(n^3)\) to \(O(n^2)\). In addition,~\cref{eq:VonNeumann_entropy} defines the Von Neumann entropy, while~\cref{eq:von_new_entropy} is an approximate calculation. It is recommended to use~\cref{eq:VonNeumann_entropy} for calculating entropy in short videos.

\subsection{Shot Segmentation}\label{ssec:shot_segmentation_of_video}

\begin{figure}[!t]
\centering
\includegraphics[width=0.6\linewidth]{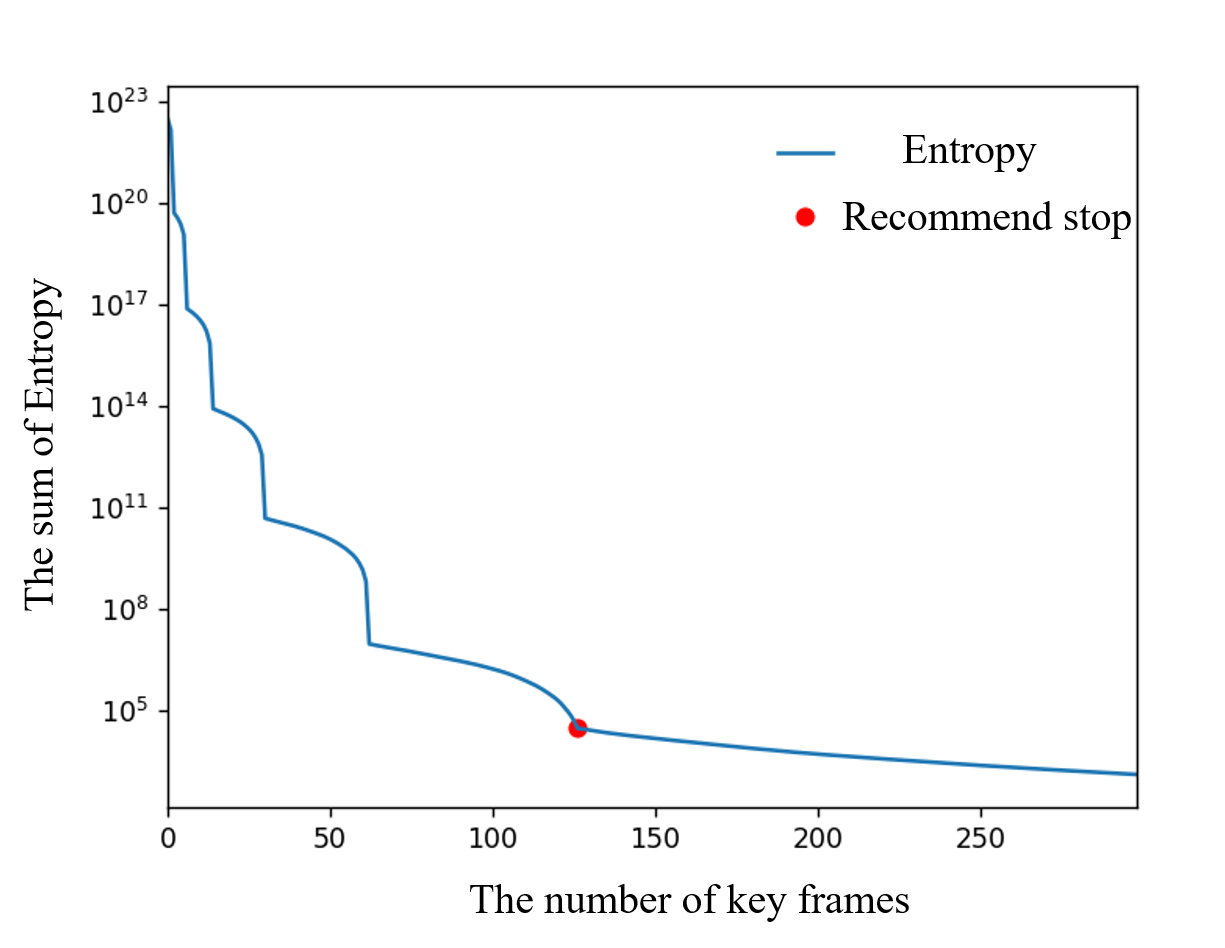}
\caption{The entropy change of a randomly selected test video to help analyze the stopping condition. The test video has 1,300 frames when sampled at 2 FPS. }
\label{fig:stop}
\end{figure}

The sum of the Von Neumann entropy of the segmented shots should be minimal. When all shots are segmented, the algorithm will continue to segment the video because not all frames in a video shot are the same. And the sum of entropy will continue to decrease at a slow rate. Therefore, when the entropy reaches the minimum, it cannot be used as a stopping condition. The entropy change caused by the intra-shot split is much smaller than that caused by the inter-shot split. When the sum of entropy decreases significantly slowly, the shot segmentation can be stopped, and this position also indicates the number of key frames. We randomly select a test video with 1,300 frames when sampled 2 FPS, and~\cref{fig:stop} shows the entropy change of the test video.~\Cref{fig:stop} shows when the number of key frames reaches 126, the entropy begins to decline significantly slowly. In fact, according to manual marking, this test video contains 120 key frames. Considering the existence of some errors in manual marking, we suggest when entropy begins to decrease significantly slowly as the stopping condition.

\begin{algorithm}[!t]
\caption{Shot segmentation}\label{alg:segmentation}
\KwIn{similarity matrix \(SimM\)}
\KwOut{shot segmentation indexes $FinalID$}
\small{\tcp{EnM saves entropy, and N represents the frame number.}}
EnM = [[-1 for n in range(N)] for n in range(N)]\\
\small{\tcp{ID saves the segmentation indexes.}}
ID = [[1, N]] \\
\small{\tcp{BS is the beam size.}}
BS = 5 \\
\small{\tcp{IDE saves segmentation indexes and their entropy.}}
IDE = []  \\
\small{\tcp{CalVN() is the function used to calculate entropy.}}
CalVN(SimM)\\
\While{not stopping condition}
{
    \For{Id in ID}
    {
        \For{id in range(1, N)}
            {\If{id in Id}{
            continue
            }
            Id.append(id)\\
            Id.sort()\\
            En = 0\small{\tcp{total entropy}}
                \For{idx in range(len(Id)-1)}
    {
        \If{EnM[Id[idx]][Id[idx+1]]==-1}   
        {EnM[Id[idx]][Id[idx+1]]=CalVN(\\SimM[Id[idx]:Id[idx+1],\\Id[idx]:Id[idx+1]])\\
        }
        En += EnM[Id[idx]][Id[idx+1]]\\
        
    }
     IDE.append([En, Id])
            }
    }
    IDE.sort()\\
    IDE = IDE[:BS]\\
    ID = [IDF[1] for IDF in IDE]\\
}
FinalID = ID[0]
\end{algorithm}

The detailed algorithm of shot segmentation is provided in~\cref{alg:segmentation}. If the shot segmentation index list is \([0, M, N-1]\), it indicates that the video is divided into two segments \([0, M-1]\) and \([M, N-1]\). The first frames of each shot are 0 and M, which are the key frames in the video.

\section{Experiments}\label{sec:experiments}

We compare our algorithm with density peak clustering (DPC)~\cite{Alex2014Clustering} to extract key frames using two datasets: the Open Video dataset and the TikTok dataset, based on two evaluation indicators: effective information rate, and redundancy rate.

\subsection{Dataset Description}\label{ssec:dataset}

In contrast to other research fields, the evaluation of key frame extraction is inherently subjective to a certain degree. The key frame extraction aims to provide humans with an intuitive understanding of video content. Therefore, existing approaches typically use human-generated key frames as the standard for comparison. The manually selected key frames are arranged in chronological order, serving as the criterion for evaluating the key frame extraction algorithm. The Open Video dataset consists of videos with varying lengths in MPEG-1 format\footnote{https://open-video.org/index.php}. Due to the time-consuming and error-prone nature of manually annotating key frames, the Open Video dataset provides us with storyboards that can be referenced for key frame extraction. Therefore, we combine the storyboard with manual annotation as the standard for key frames. We first sample the video at a certain FPS before extracting the key frames.

\subsection{Evaluation Indicators}\label{ssec:evaluation_indicator}

We use effective information rate (R), and redundancy rate (P) as the key frame extraction evaluation indicators.

\begin{figure*}[!t]
  \centering
  \begin{subfigure}{1\linewidth}
    \includegraphics[width=1\linewidth]{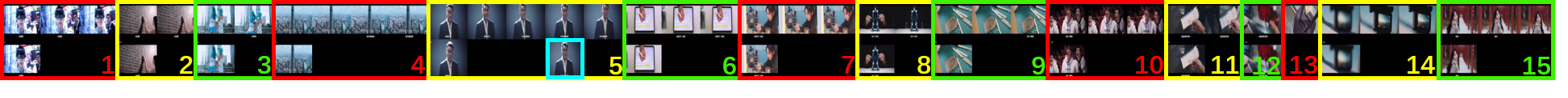}
    \caption{Key frame extraction results using our algorithm. The first row displays all frames sampled from the ``Rising Waves” fragment at 2FPS, and the second row shows the results using our algorithm. This video has 15 key frames and the blue box represents a repeated frame.}
    \label{fig:result}
  \end{subfigure}
  
  \begin{subfigure}{1\linewidth}
    \includegraphics[width=1\linewidth]{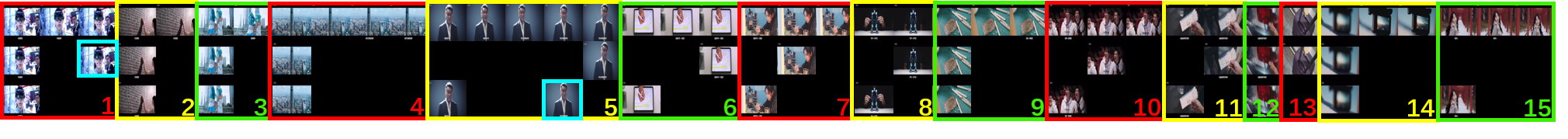}
    \caption{Comparison results between DPC and our algorithm for key frame extraction. The first row displays all frames sampled from the ``Rising Waves” fragment at 2FPS, the second row shows the results using DPC, and the last row shows the results using our algorithm. This video contains 15 key frames, and the blue boxes represent repeated frames.}
    \label{fig:compare_result}
  \end{subfigure}
  \caption{Key frame extraction results of the ``Rising Waves” fragment.}
  \label{fig:results}
\end{figure*}

\subsubsection{Effective Information Rate}\label{sssec:Effective_Rate}

The effective information rate is the ratio of the number of matched key frames to the total number of key frames, which is defined as
\begin{equation}
    \label{eq:R}
    \begin{aligned}
        R = \frac{N_{et}}{N_{gt}},
    \end{aligned}
\end{equation}
where \(N_{et}\) is the number of matched key frames and \(N_{gt}\) is the number of key frames obtained manually. R is equal to 1 when all key frames are extracted. R is between 0 and 1, and a larger R indicates that more effective information is extracted. In our key frame extraction algorithm, we select the first frame of each shot as key frames. However, the key frames provided manually may not necessarily be the first frames of the shots. Therefore, it is considered a match if the manually provided key frames are from the same shots as the extracted frames.

\subsubsection{Redundancy Rate}\label{sssec:Redundancy_Rate}

The Redundancy rate is the ratio of the number of repeated frames from extracted frames to the number of extracted frames. The redundancy rate is defined as

\begin{equation}
    \label{eq:P}
    \begin{aligned}
        P = \frac{N_{ee}}{N_{ee}+N_{et}},
    \end{aligned}
\end{equation}
where \(N_{ee}\) is the number of repeated frames from extracted frames and \(N_{et}\) is the number of matched key frames mentioned in~\cref{eq:R}.

\subsection{Numerical Experiments}\label{ssec:experimental_results}

The Open Video dataset consists of videos with varying lengths in MPEG-1 format\footnote{https://open-video.org/index.php}. The Open Video dataset provides us with storyboards that can be referenced for key frame extraction. Due to the key frame extraction is inherently subjective to a certain degree, the existing approaches typically use human-generated key frames as the standard for comparison. We combine the storyboard with manual annotation as the standard for key frames. Moreover, we first sample the video at a certain FPS before extracting the key frames.

We use a test video named ``Rising Waves” to display our results. ``Rising Waves” is a promotional video published by Bilibili in 2020. The full version lasts for 3 minutes and 52 seconds. ``Rising Waves” comprises 152 shots, with 130 of them being video clips, accounting for approximately \(86\%\) of the video. This indicates that the video clips are the primary source of the video content. In addition, the video clips are very short and visually transient. These shots change frequently, making this video a good example to demonstrate the key frame extraction results. We select a fragment whose shots frequently change, lasting 20 seconds at 25 FPS, as our input video. We choose~\cref{eq:VonNeumann_entropy} to calculate entropy and sample the video at 2 FPS. After conducting a manual analysis, we found that the video contains 15 key frames.~\Cref{fig:result} shows the key frame extraction results using our algorithm. The first row displays all frames sampled from the “Rising Waves” fragment at 2FPS, and the second row shows the results using our algorithm. The different colored rectangles in~\cref{fig:result} represent standard shot segmentation and the blue box represents a repeated frame. From~\cref{fig:result} we can observe that in our results, the 5th shots are repeated, causing the 13th shot to not be properly segmented. The effective information rate is \(93.3\%\), and the redundancy rate is \(6.7\%\). Based on these results, the proposed algorithm demonstrates good performance in key frame extraction.

\begin{table*}[!t]
\caption{The comparison between our algorithm and DPC on the Open Video dataset of key frame extraction results. The first column indicates the video name and the total frames are sampled at a certain FPS from the test videos.}
\label{tab:Comparison_results}
\centering
    {\begin{tabular}{|c|c|c|c|c|c|c|c|c|}
        \hline 
        \multicolumn{2}{|c|}{\multirow{2}*{Video}} & \multirow{2}*{Total Frames} & \multirow{2}*{Ground Truth} & \multirow{2}*{FPS} & \multicolumn{2}{c|}{DPC} & \multicolumn{2}{c|}{Ours} \\
        \cline{6-9}
        \multicolumn{2}{|c|}{~} & ~ & ~ & ~ & P & R & P & R\\
        \hline
        \hline
        \multicolumn{2}{|c|}{New Indians 02} & 60 & 14 & 2 & 0.79 & 0.21 &  0.93 & 0.07\\
        \cline{1-9}
        \multicolumn{2}{|c|}{New Indians 03} & 76 & 19 & 2 & 0.63 & 0.37 &  0.9 & 0.1\\
        \hline
        \multicolumn{2}{|c|}{NASA Anni 04} & 278 & 34 & 2 & 0.65 & 0.35 & 0.85 & 0.15\\
        \cline{1-9}
        \multicolumn{2}{|c|}{NASA Anni 03} & 304 & 47 & 2 & 0.85 & 0.15 &  0.87 & 0.13\\
        \hline
        \multirow{2}*{Apollo 11 02} & \(O(n^3)\) & \multirow{2}*{434} & \multirow{2}*{104} & \multirow{2}*{0.5} & \multirow{2}*{0.6} & \multirow{2}*{0.4} & 0.74 & 0.26\\
        \cline{8-9} \cline{2-2}
        ~ & \(O(n^2)\) & ~ & ~ & ~ & ~ & ~ & 0.74 & 0.26 \\
        \cline{1-9} 
        \multirow{2}*{NASA Anni 10} & \(O(n^3)\) & \multirow{2}*{532} & \multirow{2}*{205} & \multirow{2}*{0.5} & \multirow{2}*{0.74} & \multirow{2}*{0.26} & 0.8 & 0.2\\
        \cline{8-9} \cline{2-2}
        ~ & \(O(n^2)\) & ~ & ~ & ~ & ~ & ~ & 0.76 & 0.24\\
        \hline
    \end{tabular}}
\end{table*}

To facilitate the comparison of results, the DPC and our algorithm extract the same number of key frames.~\Cref{fig:compare_result} displays the comparison results between DPC and our algorithm for key frame extraction in the ``Rising Waves” fragment. In the comparison shown in~\cref{fig:compare_result}, it is evident that DPC has a repeated frame in the first shot and misses the key frame in the 15th shot. Meanwhile, our algorithm exhibits a repeated frame in the 5th shot and misses the key frame in the 13th shot. However, DPC requires determining the number of clusters in advance while our algorithm does not require prior knowledge of the exact number of key frames.~\Cref{tab:Comparison_results} shows further comparison between our algorithm and DPC on the Open Video dataset of key frame extraction results. The test videos are chosen from varying lengths. In~\cref{tab:Comparison_results} the first column indicates the video name, and the total frames are sampled at a certain FPS from the test videos. Additionally, the ground truth is manually analyzed from the total frames. Since the video lasts more than 10 minutes is relatively long, we use both of~\cref{eq:VonNeumann_entropy} and~\cref{eq:von_new_entropy} to calculate the entropy separately with complexity \(O(n^3)\) and \(O(n^2)\).

\begin{figure}[!t]
\centering
\includegraphics[scale=.36]{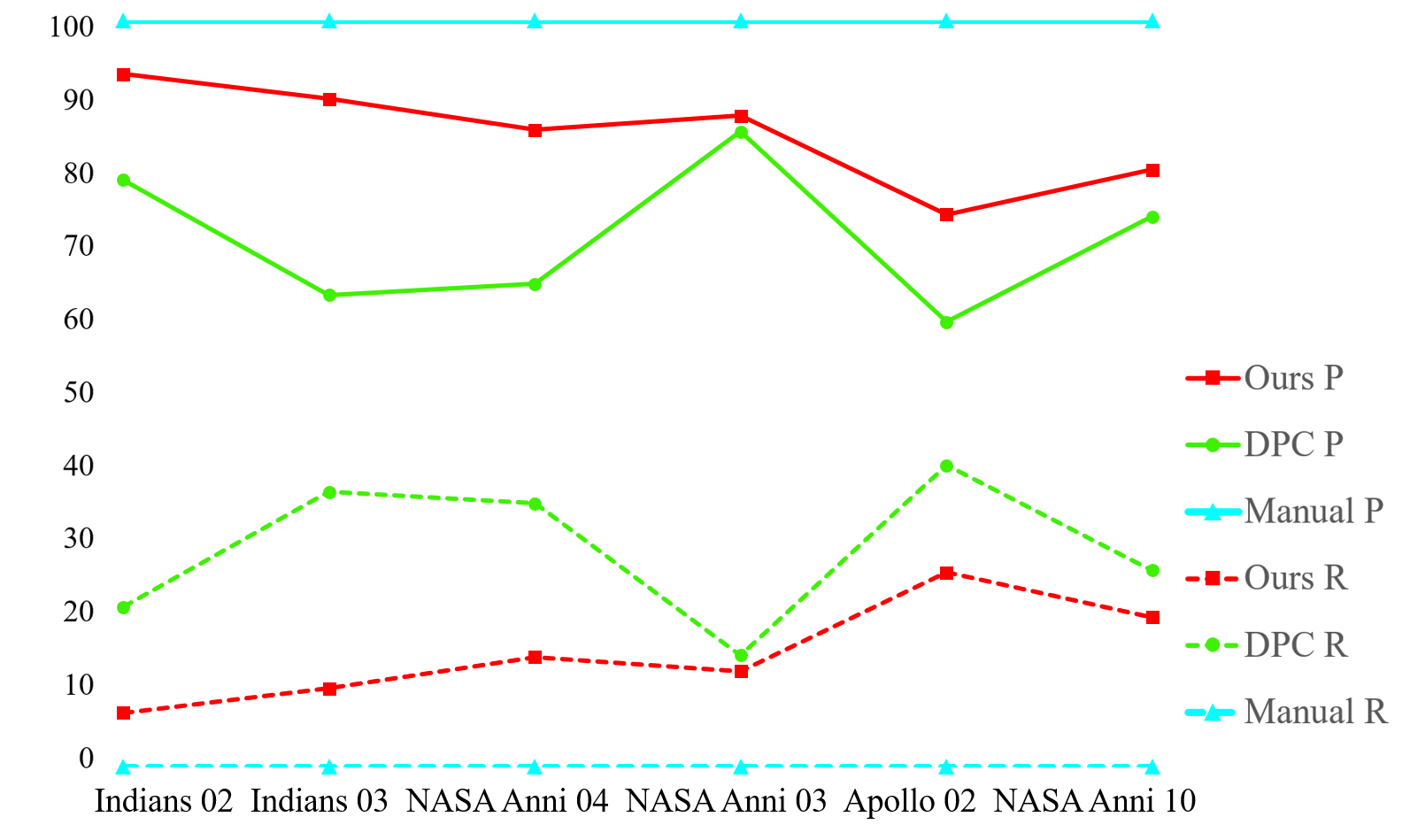}
\caption{The performance of DPC and proposed algorithm in key frame extraction on different length videos.}
\label{fig:chart_PR}
\end{figure}

From~\cref{tab:Comparison_results} we can observe that our algorithm has higher effective information rates and lower redundancy rates than DPC for varying lengths of videos. The higher effective information rate (R) indicates that the extracted key frames contain more video content. Conversely, a lower redundancy rate (P) means the extracted key frames contain less redundant information. In addition, higher R and lower P mean the extracted key frames can fully and accurately represent the original video content while minimizing redundant frames. For our algorithm, using~\cref{eq:von_new_entropy} with a complexity of \(O(n^2)\) performs slightly worse than~\cref{eq:VonNeumann_entropy} with a complexity of \(O(n^3)\) while having greater complexity. To comprehensively assess the results, we plot the distribution of P and R in~\cref{fig:chart_PR} based on the data from~\cref{tab:Comparison_results}.~\Cref{fig:chart_PR} demonstrates that the proposed algorithm exhibits greater stability for varying lengths of videos, whereas the DPC shows more fluctuation. 

\section{Conclusion}\label{sec:conclusion}

In this paper, a video key frame extraction algorithm based on shot segmentation using Von Neumann entropy has been proposed. Our approach integrates temporal information among frames, while other key frame extraction algorithms that solely focus on image features. In addition, in our approach, the number of key frames can be obtained by analyzing the change of entropy, which does not require prior knowledge of the number of key frames. We achieve this by segmenting the video into several shots and extracting the first frame from each shot as key frames. First, the video frames are sampled at specific frames per second (FPS) to reduce redundancies in the video data. Then the similarity matrix of sampled frames is calculated using the perceptual hash. After obtaining the similarity matrix, the Von Neumann entropy of the similarity matrix is computed. Then the video shots are segmented based on the Von Neumann entropy. Finally, we select the first frame of each shot as key frames. The method of shot segmentation aims to minimize the entropy of all shots, and the number of key frames is determined by analyzing the change in entropy. Our approach addresses the limitation of the clustering algorithm, which requires prior knowledge of the exact number of key frames. In addition, unlike motion analysis which can only be used in specific scenes, our algorithm is suitable for all video scenes. The curve simplification method utilizes certain image features and solely focuses on the local characteristics of the image, while our approach integrates temporal information among frames. Our approach has a computational complexity of \(O(n^2)\) which has a lower complexity than the key frame extraction algorithm based on CNNs. The experimental results show that our key frame extraction algorithm performs well in terms of video effective information rate, and redundancy rate. The extracted key frames can fully and accurately represent the video content while minimizing the number of repeated frames. Moreover, the proposed algorithm is stable and exhibits minimal fluctuation across varying lengths of videos. 


%
%
\bibliographystyle{splncs04}
\bibliography{main}
\end{document}